\pdfoutput=1

\documentclass[11pt]{article}

\usepackage{acl}

\usepackage{times}
\usepackage{latexsym}

\usepackage[T1]{fontenc}

\usepackage[utf8]{inputenc}

\usepackage{microtype}

\usepackage{titlesec}

\usepackage{times}
\usepackage{latexsym}
\usepackage{amsmath}
\usepackage{graphicx}
\usepackage{algorithm}
\usepackage{xspace}
\usepackage{amsfonts}
\usepackage{multirow}

\usepackage{amsmath,amsfonts,bm}

\def\Secref#1{Section~\ref{#1}}

\def\eqref#1{equation~\ref{#1}}
\def\Eqref#1{Equation~\ref{#1}}

\def\1{\bm{1}}

\DeclareMathAlphabet{\mathsfit}{\encodingdefault}{\sfdefault}{m}{sl}
\SetMathAlphabet{\mathsfit}{bold}{\encodingdefault}{\sfdefault}{bx}{n}

\newcommand{\E}{\mathbb{E}}

\usepackage{siunitx}
\def\sqrtexplained#1{%
  \begingroup
    \sbox0{$#1$}
    \def\underbrace##1_##2{##1}
    \sbox2{$#1$}
    \dimen0=\wd0 \advance\dimen0-\wd2
    \mathrlap{\sqrt{\phantom{\displaystyle#1}\kern\dimen0 }}
    \hphantom{\sqrt{\vphantom{\displaystyle#1}}}
  \endgroup
  #1}

\def\Doff{\mathcal{D}^\mathrm{off}}

\def\Trans{\mathcal{T}}

\usepackage{microtype}

\def\method{CALM\xspace}

\usepackage{lipsum}

\newcommand\blfootnote[1]{%
  \begingroup
  \renewcommand\thefootnote{}\footnote{#1}%
  \addtocounter{footnote}{-1}%
  \endgroup
}

\usepackage{hyperref}
\hypersetup{
    colorlinks=true,
    linkcolor=blue,
    filecolor=magenta,      
    urlcolor=cyan,
    pdftitle={Overleaf Example},
    pdfpagemode=FullScreen,
    }
\title{Context-Aware Language Modeling for Goal-Oriented Dialogue Systems}

\author{
        Charlie Snell \And
        Mengjiao Yang \And
        Justin Fu \\
        UC Berkeley \\
        \texttt{\{csnell22,sherryy,justinfu,suyi,svlevine\}@berkeley.edu}
        \\\And
        Yi Su \And
        Sergey Levine
  }

\date{}

\begin{document}
\maketitle
\begin{abstract}
Goal-oriented dialogue systems face a trade-off between fluent language generation and task-specific control. While supervised learning with large language models is capable of producing realistic text, how to steer such responses towards completing a specific task without sacrificing language quality remains an open question. In this work, we formulate goal-oriented dialogue as a partially observed Markov decision process, interpreting the language model as a representation of both the dynamics and the policy. This view allows us to extend techniques from learning-based control, such as task relabeling, to derive a simple and effective method to finetune language models in a goal-aware way, leading to significantly improved task performance. We additionally introduce a number of training strategies that serve to better focus the model on the task at hand. We evaluate our method, Context-Aware Language Models (\method), on a practical flight-booking task using AirDialogue.
Empirically, \method outperforms the state-of-the-art method by 7\% in terms of task success, matching human-level task performance.
\end{abstract}

\setlength{\abovedisplayskip}{1pt}
\setlength{\abovedisplayshortskip}{1pt}
\setlength{\belowdisplayskip}{1pt}
\setlength{\belowdisplayshortskip}{1pt}
\setlength{\jot}{1pt}
\setlength{\floatsep}{1ex}
\setlength{\textfloatsep}{1ex}
\setlength{\parskip}{0em}
\titlespacing\section{0pt}{5pt plus 1pt minus 1pt}{0pt plus 1pt minus 1pt}
\titlespacing\subsection{0pt}{5pt plus 1pt minus 1pt}{0pt plus 1pt minus 1pt}
\titlespacing{\paragraph}{0pt}{2pt}{0.5em}

\section{Introduction}

\blfootnote{Code at \url{https://sea-snell.github.io/CALM_LM_site/}}

Dialogue systems have typically approached the problem of generating realistic dialogue from the perspective of supervised learning~\citep{duvsek2016sequence,eric2017copy,mei2017coherent,chen2019semantically,wu2019alternating,hosseini2020simple,peng2020soloist,adiwardana2020towards}. However, dialogue can also be viewed as a sequential decision making process, which is well-suited to planning and reinforcement learning (RL) algorithms. A challenge with the classical RL approach to dialogue is the requirement for active interaction with humans~\citep{gavsic2011online}. Training such a system with active human-in-the-loop interaction quickly becomes expensive and cumbersome, making it desirable to develop techniques for goal-directed training of dialogue systems that can effectively leverage offline data.

\begin{figure}
    \centering
    \includegraphics[scale=0.4]{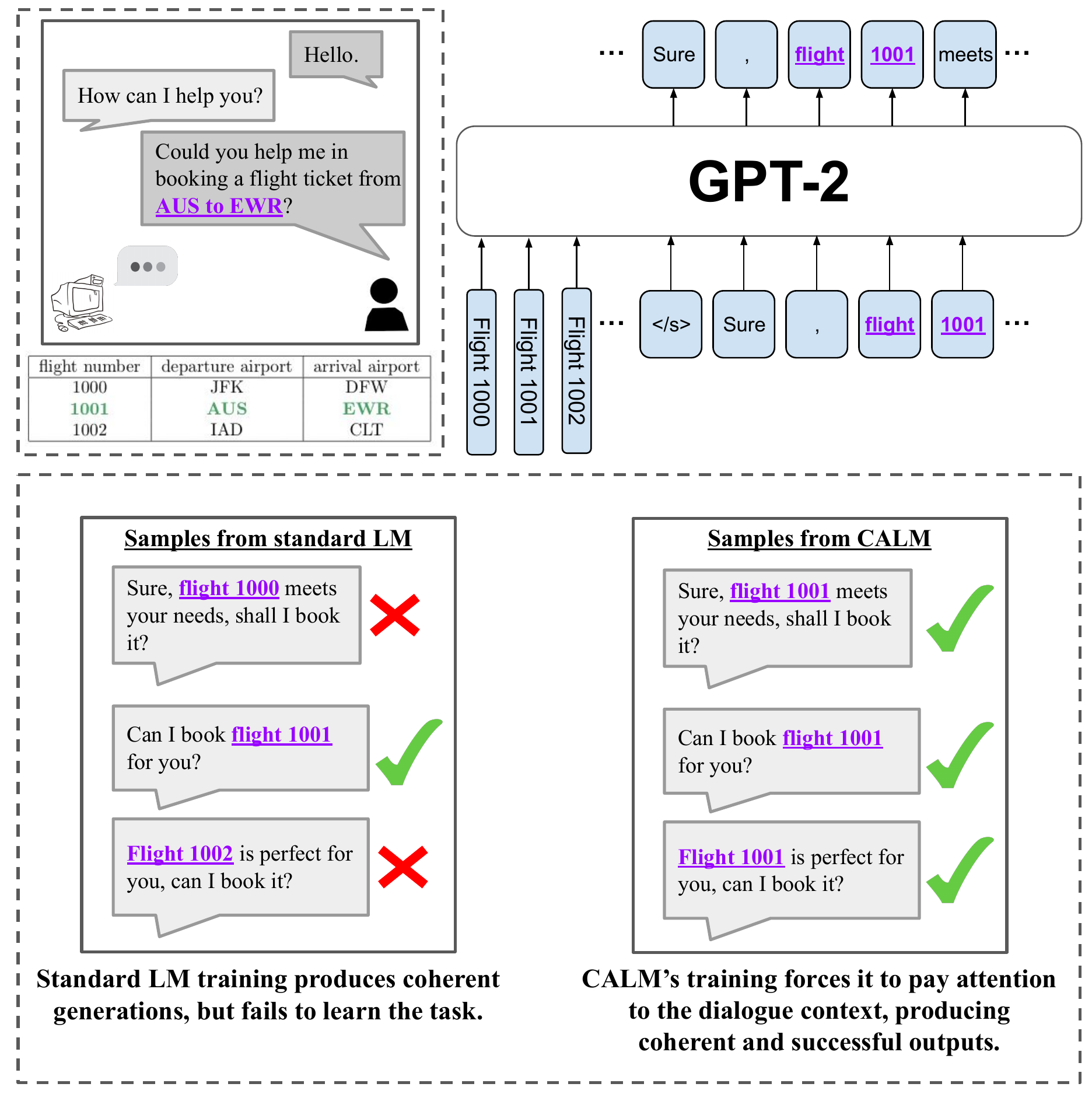}
    \caption{\textbf{\method is an end-to-end language model for goal oriented dialogue.} \method's training objective teaches the model to better pay attention to the dialogue task context, yielding a $\sim$50\% improvement in task success over standard LM training on a flight booking task.}
    \label{fig:teaser}
\end{figure}

While many dialogue generation techniques based on RL and learned control have been proposed~\citep{eckert1997user,levin2000stochastic,chung2004developing,georgila2006user,schatzmann2007agenda,heeman2009representing,georgila2011reinforcement}, most such systems take a pipelined approach, where an abstract representation of states and actions is designed by hand and then combined with RL to train a ``dialogue management'' system, rather than generating dialogue end-to-end. These pipelined approaches rely on a manually designed decomposition of the dialogue task, which may be domain-specific and, more importantly, may not enjoy all of the benefits of tightly integrating low-level text generation with the overall goals of the task.
In this work, we instead ask: how can we scalably and effectively introduce the mechanisms of goal-directed decision making into end-to-end language models, to directly steer language generation toward completing specific dialogue tasks rather than simply generating probable responses?

\begin{figure*}
    \centering
    \includegraphics[scale=0.38]{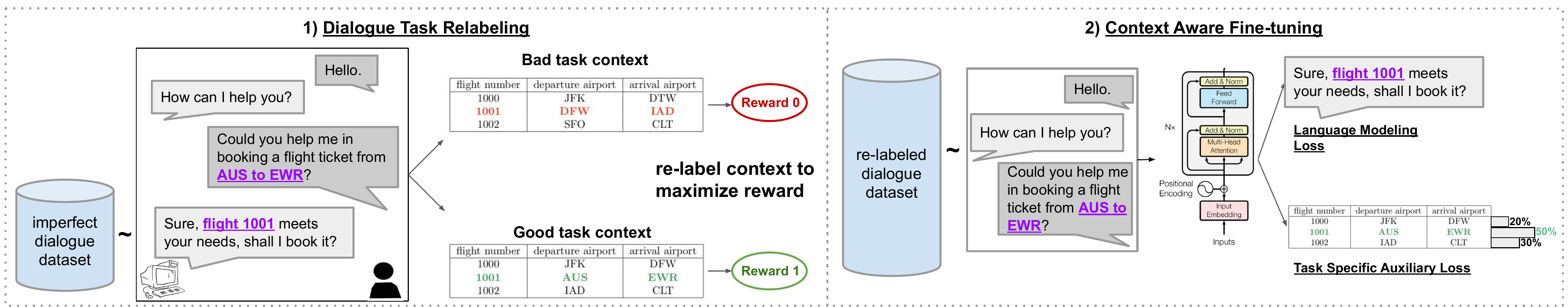}
    \caption{\textbf{A visual outline of \method.} We apply Task Relabeling to our static offline dataset, by swapping out the task context –- in this case a flight table –- such that the attached dialogue becomes an optimal example of task completion. When fine-tuning on this relabeled dataset, we then apply a Task Specific Auxiliary Loss on top of the standard language modeling objective; this helps the model learn to use the task context. Once trained, \method can consistently solve goal-directed dialogue tasks.}
    \label{fig:main}
\end{figure*}

To this end, rather than utilizing a pipelined approach, we aim to directly finetune language models in a task-aware manner such that they can maximize a given utility function. We observe that large language models can already be formulated within a Markov decision processes (MDP) as capturing both the dynamics and policy for a decision-making task, where dialogue history serves as state, and the agent's utterances serve as actions. 
We could utilize this observation by finetuning the models directly with online RL, but the need for human-in-the-loop training makes this difficult. Offline RL methods~\citep{levine2020offline,fujimoto2019off,wu2019behavior,wang2020critic} provide an alternative approach, but typically require value function estimation, which is not straightforward to perform with a language model.
Instead, we propose a conditional imitation learning strategy coupled with a novel task relabeling approach that can finetune language models from offline data, such that the model still represents the joint distribution over dialogues, but \emph{tilts} this distribution toward dialogues with a high reward. This amounts to a \emph{task-aware} finetuning strategy that integrates task information into the model.
The main contribution of our work is \method (\emph{Context-Aware Language Modeling}), a framework for end-to-end goal-directed dialogue generation.
\method unifies the traditional language modeling objective with task-specific supervision, where a language model is interpreted as a joint representation of dynamics and policies in an MDP, and the finetuning process utilizes a conditional imitation learning objective with a novel task relabeling strategy that teaches the model how to generate high-utility dialogues (see Figures \ref{fig:teaser} and \ref{fig:main}).
Because \method interprets the language model as \emph{both} a dynamics model and a policy, it can be used as either a model-free method, where the dynamics are discarded and the policy component is used to greedily generate responses, or as a \emph{model-based} method, where the dynamics component can be used to \emph{plan} at test-time.
We empirically evaluate \method on AirDialogue~\citep{wei2018airdialogue}, the largest dataset for goal-oriented dialogue based-on a flight-booking task. \method improves the task success by 10\% over the previous state-of-the-art method~\citep{chen2020airconcierge} following the evaluation protocol proposed by~\citet{wei2018airdialogue}, achieving the first-ever human-level performance on this dataset.

\section{Related Work}

Our goal is to enable end-to-end training of goal-directed dialogue agents. In these settings, an agent aims to complete a particular task with its utterances~\citep{smith1994spoken}. Goal-directed agents have been explored in contexts such as personal assistants~\citep{mctear2002spoken,budzianowski2018multiwoz, williams2014dialog}, recommendation systems~\citep{liu2010dialogue,kang2019recommendation}, education~\citep{yuan2008human}, and negotiation~\citep{he2018decoupling,lewis2017deal}.
While there are multiple approaches to constructing dialogue agents, in this work we frame the problem of generating dialogue as a sequential decision making problem within a (partially observed) Markov Decision Process (MDP)~\citep{singh1999reinforcement,young2013pomdp}. Prior works that utilize such an MDP formulation typically aim to train a dialogue management system~\citep{singh2002optimizing}, in which the agent reasons about higher-level abstractions of the state of the conversation, and language generation is performed using a downstream procedure. Dialogue management systems have been trained using techniques such as online reinforcement learning via policy gradients~\citep{gavsic2011online,he2018decoupling}, off-policy reinforcement learning~\citep{pietquin2011sample,yu2016strategy} or actor-critic methods~\citep{su2017sample}. Our method differs from dialogue management systems in that \method is an end-to-end system optimized for successful task completion, and performs both high-level decision making and language generation.

Recent advancements in language models, such as recurrent neural networks~\citep{sundermeyer2012lstm,asri2016sequence,su2016continuously,zhao2019rethinking,wang2020modelling,zhang2020task} and attention-based architectures~\citep{vaswani2017attention,liu2019roberta,devlin2018bert,brown2020language}, have spurred increasing interest in such end-to-end dialogue systems~\citep{hosseini2020simple,peng2020soloist,adiwardana2020towards}. Model-based approaches, in which a learned agent is substituted for a human, allow learning to be done entirely within simulation without human intervention~\citep{li2016deep,he2018decoupling,kang2019recommendation,lewis2017deal,liu2018dialogue}. In contrast to these approaches, \method augments the traditional language modeling objective with task-specific rewards in order to finetune a model that is more aware of task goals, which significantly improves performance over a na\"{i}ve language model without the need for simulating human responses in an interactive training loop. \citet{jaques2019way} recently proposed a model-free, offline approach to undirected dialogue, or dialogue without a specific task goal. Our method differs in that we aim to solve goal-oriented dialogue which allows us to optimize task-specific objectives, and that we take a model-based RL approach which enables us to leverage fine-tuned language models.

\section{Preliminaries}
\label{sec:background}
In this section, we review our notation and problem formulation for casting dialogue within a sequential decision making framework.

\paragraph{POMDP formulation.}
We formulate dialogue generation as a partially observable Markov decision process (POMDP)~\citep{kaelbling1998planning}, with a state that consists of known and unknown context information about the task.
Let $c_h\in\mathcal{C}^{(h)}$ denote the hidden context for the task, and let $c_o\in\mathcal{C}^{(o)}$ denote the observed context. For instance, in a flight booking task, a table of available flights might correspond to $c_o$, while the particular flight that the human wants to book, which is unknown to the agent, corresponds to $c_h$.
Note that the reward, which requires booking the right flight, depends on both hidden and observed contexts. We can define such an environment as a POMDP $\mathcal{M} = (\mathcal{S}, \mathcal{A}, \mathcal{O}, \mathcal{T}, \mathcal{Z}, \mu_0, \mathcal{R}, \gamma)$. We denote a conversation $\tau$ as $\tau:=\{a_0,e_0,...,a_{T}\}$, where $T$ denotes the number of turns in a conversation and $a_t$ and $e_t$ represent utterances (strings of tokens) from the dialogue agent ($a_t$) and the human ($e_t$) at the $t$-th turn, respectively. We additionally use $\tau_{<t}$ to denote conversation history up to the $t$-th turn.
We can represent the underlying POMDP state $s_t \in \mathcal{S}$ as the concatenation of both of the contexts and the previous conversation history $s_t := \{c_h, c_o, \tau_{<t}\} = \{c_h, c_o, a_0, e_0,..., a_{t-1}, e_{t-1}\}$. However, we only observe the last two elements of the state tuple, such that our observation $o_t\in \mathcal{O}$ at the $t$-th conversation turn is $o_t= \{c_o, \tau_{<t}\}$.
An action $a_t \in \mathcal{A}$ is the agent's response to the current state $s_t$. Given our definition of the state, the full conversation in a dialogue can be conveniently represented by the last observation and action, $\{o_T,a_T\}$. An agent $\pi: \mathcal{O}\to\mathcal{P}(\mathcal{A})$ maps observations to sets of probability measures over the action space $\mathcal{P}(\cdot)$. A transition function $\mathcal{T}(\cdot|s_t, a_t)$, represents a distribution over the human's utterances, returning $s_{t+1}$ as the state at turn $t+1$.
We only consider the sparse reward setting with $r_T=R(s_T,a_T)\in\{0, 1\}$ denoting task completion, and $r_t=0$, $\forall t<T$. Our final reward is therefore dependent on both the context and the dialogue: $R(s_T,a_T)=R(\tau, c_h, c_o)$, where the context $\{c_o, c_h\}$ is randomly sampled for each dialogue from some initial distribution $\mu_0$.

\paragraph{Goal-oriented dialogue.}
Goal-oriented dialogue systems aim to maximize the expected reward of the above POMDP
\begin{equation}
\label{eq:rl-obj}
    \textstyle
	\E_{\{c_o, c_h\} \sim \mu_0,\pi,\mathcal{T}} [\sum_{t=0}^{T}\gamma^t R(s_t,a_t)],
\end{equation}
where $\{c_h, c_o\}$ is sampled from distribution $\mu_0$. \textit{On-policy} RL algorithms optimize this objective via environment interaction, which is represented by a real human.
However, because human-in-the-loop training is expensive, we pursue an \emph{offline learning} approach where we are given a fixed dataset and there is no further interaction with the human in the learning process. This dataset is composed of $n$ trajectories with $\Doff = \{c_h^{(i)}, c_o^{(i)}, \tau^{(i)}, r^{(i)}\}_{i=1}^n$ 
with each $\tau^{(i)}=\{a^{(i)}_0,e^{(i)}_0,,...,a^{(i)}_T\}$ and its corresponding final reward for task completion $r^{(i)}$. Our goal is to learn the policy $\pi(a|o)$ which improves the dialog agent's ability in achieving the highest task reward defined in~\Eqref{eq:rl-obj}.

\paragraph{Language models.}
While conventionally a language model is seen simply as a sequence model over tokens of the form $\prod_{t=1}^T p(x_{t+1} | x_{1:t})$, when the sequence $x_{1:T}$ corresponds to a dialogue trajectory $\tau$, we can also interpret a language model as learning the distribution over $\tau$. This distribution can be factored into the product of the policy $\pi(a_t | \tau_{< t})$ and the dynamics $\Trans(\tau_{<t+1}|\tau_{<t}, a_t)$, and so we can say that a language model \emph{also} represents the policy and the dynamics. Therefore, the maximum likelihood objective for training or finetuning a language model on a dialogue dataset $\Doff$ consisting of dialogue trajectories $\tau$ can be written as
\begin{align}
    \label{eq:lm-obj}
    \mathcal{L}_{LM}(\theta) =& \max_{\theta} \mathop{\mathbf{E}}_{\tau \sim \Doff} \sum_{t=1}^{T}  \bigg(\log{\pi_{\theta}(a_t|\tau_{<t})} \nonumber\\ &+ \log{\Trans_{\theta}(\tau_{<t+1}|\tau_{<t}, a_t)}\bigg),
\end{align}
where $\pi_{\theta}(a_t|o_t)$ represents a policy that generates new dialogue based on the observed context and dialogue history, and $\Trans_{\theta}(\tau_{<t+1}|\tau_{<t}, a_t)$ represents the observed dynamics characterizing human responses, and $\theta$ denotes parameters in $\pi$ and $\Trans$. Note that $\tau_{<t}$ consists only of the conversation history, and does not contain any task-specific context.
A na{\"i}ve approach to train dialogue systems is to jointly parameterize both $\pi$ and $\Trans$ as one language model, and optimize~\Eqref{eq:lm-obj} on pre-collected conversations $\Doff$. This method corresponds to behavioral cloning (BC)~\citep{pomerleau1989alvinn}.

\paragraph{Context conditioning.} While an agent trained using~\Eqref{eq:lm-obj} can learn policies and dynamics that imitate human conversations, this objective does not incorporate the task goal, and may not produce a policy that is more performant than the dataset $\Doff$. While it is possible to input $c_o$ into the language model to maximize the conditional probability of $P(\tau|c_o)$ using a conditional version of the language modeling objective, $\mathcal{L}_{CTX}(\theta)$,
\begin{align}
    \label{eq:context-lm-obj}
    \mathcal{L}_{CTX}(\theta) =& \max_{\theta} \mathop{\mathbf{E}}_{(\tau, c_o) \sim \Doff} \sum_{t=1}^{T} \Bigl(\log{\pi_{\theta}(a_t|\tau_{<t}, c_o)} \nonumber\\ &+ \log{\Trans_{\theta}(o_{t+1}|\tau_{<t}, a_t, c_o)}\Bigr),
\end{align}

contexts with particular task structures (e.g., a set of entries in a table) may not be simply processed as a sequence similarly to $\tau$. Additionally, the language model is not pretrained to read structured context, and oftentimes the recent dialogue history is much more predictive of the next utterance than the task context is. As a result, language models can ignore the task context and only learn $P(\tau)$ despite being conditioned on $c_o$. Our approach builds on this conditional modeling approach, but makes a number of improvements that allow it to be more aware of the context information, which attains significantly better results in our experiments.

\section{Context-Aware Language Modeling}

In this section, we present our method for goal-oriented dialogue systems, Context-Aware Language Modeling (\method).
\method interprets a language model as a combination of a policy and a dynamics model in the POMDP formulation of a dialogue task, as described in \Secref{sec:background}. Under this interpretation, na\"{i}ve supervised finetuning on the dialogue dataset can be viewed as behavioral cloning (BC)~\citep{pomerleau1989alvinn}. However, BC only imitates data and does not necessarily produce a \emph{good} policy in terms of completing tasks. We propose to improve the policy by utilizing a task relabeling strategy (described in~\Secref{sec:relabel}), analogous to prior task relabeling approaches~\citep{kaelbling1993learning,andrychowicz2017hindsight,pong2018temporal,savinov2018semi,ghosh2019learning,lynch2020learning,eysenbach2020rewriting}. 
This relabeling procedure augments the data with examples of near-optimal utterances, making the language model more task-aware.
However, we find several shortcomings with this approach alone and propose the following improvements. First, an expressive language model is liable to ignore the task context, which we address by proposing an auxiliary loss (\Secref{sec:auxiliary}) that forces the model to utilize this information. Second, learning from structured task information is difficult and can result in models that fail to capture complex task structure, so we propose a task pre-training procedure to improve the learnability (\Secref{sec:pretrain}). Finally, to further improve performance we use a model-based planning procedure (\Secref{sec:rollouts}) on top of the proposed method that samples multiple dialogues in parallel and selects the most promising candidates. 

\subsection{Dialogue Task Relabeling}
\label{sec:relabel}
$\mathcal{L}_{CTX}(\theta)$
defines a context-conditional maximum likelihood objective for training an expert imitation policy in conjunction with a dynamics model.
However, simply imitating all the dialogue data does not necessarily produce the best possible policy. We would like to learn a policy that produces dialogue that is more optimal, in the sense of better maximizing the task utility, than the average dialogue in the dataset.
Task relabeling enables us to learn from optimal trajectories without simply filtering the dataset for high-reward trajectories, which would unnecessarily discard potentially informative data.
In the case of dialogue, we can perform task relabeling by considering the context $\{c_o, c_h\}$ as defining the task. While a given dialogue may be unsuccessful for the context for which it was collected, it could be considered successful under a different context. In this case, we can simply swap out $\{c_o, c_h\}$ to create optimal task examples from the many sub-optimal examples provided by $\Doff$. Since our reward $R(c_h, c_o, \tau)$ is a function of the dialogue and context, we can modify the reward for a given dialogue just by changing the given observed context $c_o$. Using this observation, we can relabel unsuccessful dialogues with successful ones, and even for already successful dialogues there may be multiple $c_o$ corresponding to task success, allowing us to augment the number of successful $(c_h, c_o, \tau)$ tuples.

Formally, since our POMDP includes a prior distribution over contexts $\{c_h, c_o\}\sim\mu_0$, there exists a posterior $q(c_o|\tau, c_h)$ over observed contexts that correspond to optimal task completion under a given $\tau$. We can then re-label $\tau$ to be optimal under its context by sampling a new $c_o$ from $q(c_o|\tau, c_h)$. In practice, this sampling is performed by rejection sampling from either $\mu_0$ or some $P(c_o|c_h)$; the latter, lower entropy distribution, can be preferred if there is a low probability of sampling valid, high-reward contexts under $\mu_0$. Now, given any $\tau$ from an offline dataset of dialogues, we can learn from the full distribution of contexts corresponding to optimal task completion under this dialogue.

In order for this relabeling procedure not to bias our policy towards behavior that is overly-optimistic about the user's responses, it is necessary that the distribution of these responses in our dataset does not depend on the portion of the context that is relabeled. For example, relabeling the table of available flights for a flight booking task should generally be reasonable, because the user is usually unaware of the flight table. On the other hand, relabeling the desired flight would not make sense, since the user's utterances are strongly depend on this. To provide another example, in a bargaining task~\citep{lewis2017deal}, the agent might fail to obtain the desired item and instead get an item of lesser value. But relabeling with a context that assigns a higher value to the item received would not lead to a reasonable example, since the agent mainly received this item as a result of the user's responses rather than as a result of their own bargaining skill.

Methods based on similar principles have previously been proposed in the deep RL community for simple parametric tasks, such as goal-reaching or linearly-parameterized reward functions~\citep{kaelbling1993learning,andrychowicz2017hindsight,eysenbach2020rewriting}. However, the dialogue task relabeling that we employ is particularly effective in our setting, since there may be exponentially many contexts that are optimal for a given dialogue (e.g., many different flight tables for a flight booking task), in contrast to the simpler task parameterizations used in prior work, where for example only one goal might be optimal for a given trajectory (the one that is reached). As a result, this technique not only allows us to turn sub-optimal task data into optimal data, but it also allows us to greatly increase the number of optimal task examples from which we can learn, which we will show leads to a large performance improvement.

\subsection{Task-Specific Auxiliary Loss}
\label{sec:auxiliary}

Goal-oriented dialogue generation can be viewed as learning the conditional distribution $P(\tau|c_o)$, where $\tau$ represents the generated dialogue given a specific context $c_o$. However when trained na\"{i}vely, language models are liable to ignore this conditioning context, instead focusing purely on the previous utterances in the dialogue.
In this case, the model is effectively only learning $P(\tau)$ despite having both the capacity and the context to learn the lower-entropy conditional distribution $P(\tau|c_o)$.

While dialogue tasks are by definition carried out through natural language, there is often an abstract high-level action $\alpha_h \in \mathbf{A}$ that essentially determines the success of the task. In the case of the information retrieval task that we consider in this paper, these high-level actions correspond to deciding which database entity to retrieve for the user (e.g., suggesting a flight to the customer that meets all of their needs). While these high-level actions are theoretically learnable from correlations between the dialogue and the given context, in general, we find that learning these correlations corresponds to a relatively small decrease in dialogue entropy under the model. As a result, the model is less incentivized to learn these correlations relevant to the task than the form of the dialogue. To address this issue, we incorporate an auxiliary objective into our training, which trains the model directly to predict the abstract high-level actions taken in the present dialogue. This objective effectively up-weights gradients relevant for learning the high-level actions, which further helps the model to utilize the context to solve the high-level task through dialogue.

For a given dialogue-context pair $(\tau, \{c_h, c_o\})$ and high-level action, $\alpha_h$, our auxiliary objective is then simply to maximize the likelihood of the high-level actions taken in the dialogue:

\begin{equation}
    \mathcal{C}(\phi) = \max_{\phi} \mathop{\mathbf{E}}_{(c_h, c_o, \tau, \alpha_h) \sim \Doff} \log{P_{\phi}(\alpha_h|\tau, c_o)}.
\end{equation}

Just like the language modeling objective, this classification objective is averaged over each token in the dialogue sequence. Our full training objective then becomes:
\begin{equation}
    \max_{\theta, \phi} \mathcal{L}_{CTX}(\theta) + \beta * \mathcal{C}(\phi),
\end{equation}
where $\beta$ is a hyper-parameter and $\mathcal{L}_{CTX}(\theta)$ is the standard context-conditional language modeling objective as defined in Section \ref{sec:background}.

\subsection{Task Pretraining}
\label{sec:pretrain}
As observed by \citet{liu2021tapex}, for some structured tasks, such as table question answering, pre-training on a simplified version of the given task with a synthetic context can help the model to focus learning on the ``skills'' that are most relevant to utilize the task context, which leads to improved downstream task performance. We instantiate this idea in our method by pre-training our model on a simplified (dialogue-free) version of the task. 
Instead of simultaneously modeling all the details of the raw dialogue, as is required to learn $P(\tau|c_o)$, the key observation here is that in our case the task reward only depends on the tuple $\{c_h, c_o, a_T\}$. This enables us to effectively learn to execute the task by only modeling $P(c_h, a_T|c_o)$, without any dialogue at all. 
By pre-training our model to first learn this simplified distribution, we effectively focus on learning the necessary skills for completing the task. It is expected that the skills learned during this pre-training phase should also generalize and transfer when we later perform training on the real dialogue. The particular instantiation of this principle in the case of AirDialogue is described in Section~\ref{app:pretrain}.

\subsection{Model-Based Dialogue Rollouts}
\label{sec:rollouts}

While the methodology discussed so far can produce effective policies, language models also represent task dynamics, as discussed in Section~\ref{sec:background}. We can leverage this fact to further improve the performance of our fine-tuned models by performing model-based planning at test-time, using both the policy and dynamics components in concert to further maximize task reward. A full dialogue trajectory can then be formed by concatenating this sampled future trajectory $\tau_{\geq t}$ with the current state of the dialogue $\tau_{<t}$ i.e., $\tau=\{\tau_{<t}, \tau_{\geq t}\}$. We perform the model-based planning by sampling $k$ such future trajectories from the final fine-tuned model, and ranking them according to an estimated reward function $\hat{R}(\tau, c_o)$ (see Appendix \ref{app:rew}). Then, we improve upon the policy $\pi$ from which we took the samples by taking the action (i.e., the next utterance) $a_t$ which receives the highest estimated reward among the sampled trajectories. This rollout sampling procedure is identical to the one used by \citet{lewis2017deal}.

\section{\method for AirDialogue}

In this section, we instantiate our proposed method, \method, for the AirDialogue flight booking task \citep{wei2018airdialogue}. We first give an overview of the task, and then describe how to do relabeling and context conditioning on this specific task.

\subsection{AirDialogue Dataset}

\paragraph{Dataset overview.}
The AirDialogue dataset \citep{wei2018airdialogue} is a recently published large-scale airline reservation dataset based on the aforementioned task. The dataset includes 402,038 conversations. The dataset involves three distinct tasks: booking, canceling, and changing flights. We describe the booking task in detail below.

\paragraph{Flight booking task.}
The (human) customer is given a set of 12 trip requirements, and the flight agent (bot) is provided with a table of 30 flights. The goal of the flight agent is to book a flight from the table for the customer which meets all their requirements, or to correctly inform them that no such flight is available. To determine task success, the flight agent must predict an explicit action at the end of the dialogue indicating the flight that was booked or inform no flight available. See Figure~\ref{fig:AirDialogueExample} for an example conversation from the dataset.

\subsection{Processing Tables}
The AirDialogue booking tasks require efficiently querying a flight table containing flight information (e.g., departing location, ticket price) given to the agent prior to the conversation. In order to successfully complete the booking task, the agent needs to be able to filter, select, and integrate information from the flight table based on the customer's preferences inferred from the dialogue.

Instead of treating the tables as unstructured sequences~\citep{wei2018airdialogue, jiang2021towards} or as SQL databases~\citep{chen2020airconcierge}, \method models tables as an observable context consisting of a set $c_o = \{f_1, f_2, f_3, ..., f_N\}$ of table rows. These rows are then input to our model as a set of embeddings (see appendix \ref{app:fl-bot} and \ref{app:table} for more details).

\subsection{Relabeling AirDialogue with \method}
While the AirDialogue dataset only includes one flight table for each dialogue, there are potentially many flight tables compatible with each dialogue as each flight can appear in many tables. We hence implement our relabeling procedure as described in Section \ref{sec:relabel} as follows. We perform rejection sampling on the observable context (i.e., the table of flights) $c_o\sim q(c_o|\tau, c_h)$, sampling until we obtain a new context $(c_h, c_o, \tau)$, which gives maximum reward possible $R(\tau, c_h, c_o)=\text{max}_{c_o}R(\tau, c_h, c_o)$.
The prior distributions $p(c_o)$ and $p(c_o|c_h)$, from which the tables in the AirDialogue dataset were sampled, are provided with the dataset. By rejection sampling from $p(c_o|c_h)$, we can effectively sample from the posterior $q(c_o|\tau, c_h)$ within a certain computational budget. In this setting, $c_o$ denotes tables and there are exponentially many tables which correspond to a task success under a given dialogue. Therefore, with our relabeling approach, we increase the number of near-optimal task examples exponentially, which makes it much easier for the language model to learn to query the flight table.

Our relabeling is approximately valid according to the condition specified in Section \ref{sec:relabel}. While the customer does not have access to the flight table and therefore is not directly affected by our relabeling, there are still some minor edge-cases in which over-optimism about the dynamics could be learned by our policy. If for example, in the dataset the customer were to occasionally reject the first flight that we suggest, our policy may learn to assign a small probability to the action of initially offering the wrong flight, relying on them subsequently rejecting it such that we can later recover and offer the correct one. However, in practice we observe that these cases are rare in AirDialogue.

\subsection{Table Selection as Auxiliary Loss}
The primary high-level action involved in AirDialogue is the decision of which flight table entry, if any, to recommend to the user. We therefore implement our auxiliary objective as a classification head on top of the language model, trained to predict the flight table entry
that meets the customer's requests. Specifically, our set of high-level actions $\mathbf{A}$ is the set of flight table rows $\{f_1, f_2, f_3, ..., f_N\}$ plus an additional item $f_0$, corresponding to the case in which no flights meet the customer's requirements. If $f^*$ is the flight recommended in the dialogue, then our auxiliary objective is:

\begin{equation}
    \mathcal{C}(\phi) = \max_{\phi} \mathop{\mathbf{E}}_{(c_o, \tau) \sim \Doff} \log{P_{\phi}(f^*|\tau, c_o)}.
\end{equation}

\section{Experiments}

\begin{figure*}
    \centering
    \includegraphics[scale=0.325]{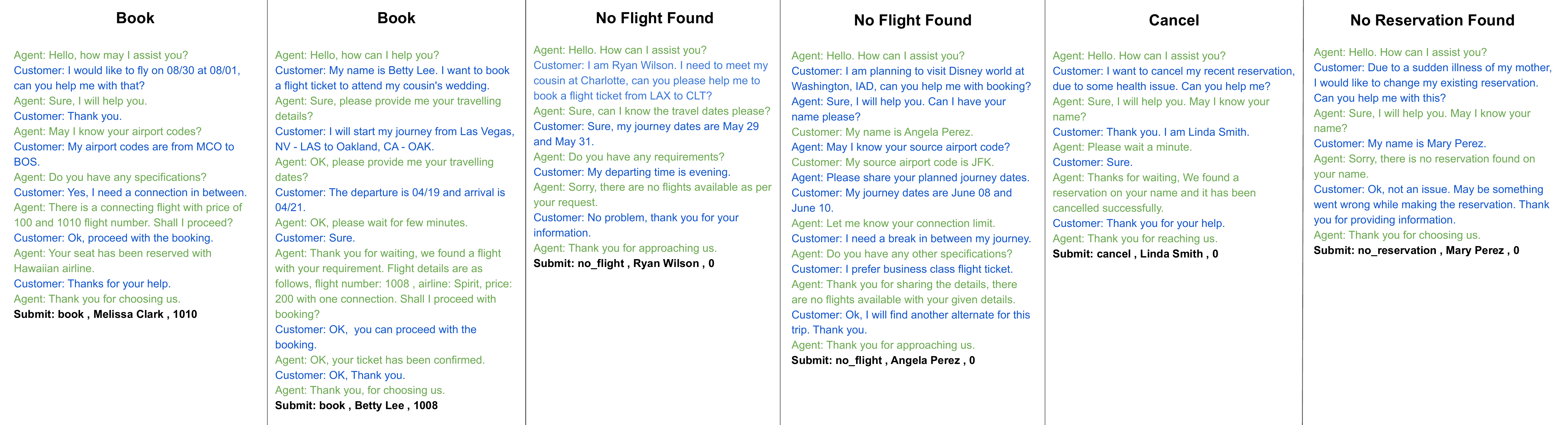}
    \caption{\textbf{Example dialogues generated by \method (in green) in the simulated evaluation.} Despite being end-to-end, \method produces highly coherent and sensible outputs.}
    \label{fig:DialogueExamples}
\end{figure*}

\begin{table}[]
    \footnotesize
    \centering
    \begin{tabular}{|l|c|}
        \hline
        & success rate \\
        \hline
        \method (greedy) & \textbf{0.88} $\pm$ 4e-3 \\
        LM(GPT2-small) (greedy) & 0.38 $\pm$ 1e-3 \\
        AirConcierge (greedy) & 0.81 $\pm$ 7e-3 \\
        \method (planning) & \textbf{0.90} $\pm$ 2e-3 \\
        LM(GPT2-small) (planning) & 0.74 $\pm$ 7e-3 \\
        Human & \textbf{0.88} \\
        \hline
    \end{tabular}
    \caption{\textbf{Comparison of our method and baselines across all tasks.} Using greedy decoding, our method matches human performance, greatly improving over baselines. Adding roll-outs (32 samples) further improves task completion.}
    \label{tab:main_eval}
\end{table}

\label{sec:exp}
In this section, we empirically evaluate the performance of \method on AirDialogue~\citep{wei2018airdialogue}. We first show that \method outperforms the SOTA on the AirDialogue dataset by around $7\%$ in the standard simulated evaluation protocol proposed by \citet{chen2020airconcierge}, which prior work denotes as ``self-play" (see Appendix~\ref{app:self-play}), and this matches human-level performance as reported by \citet{wei2018airdialogue}.
Beyond this, we also perform a comprehensive set of ablation studies to validate the necessity of each component of \method.

\paragraph{Experiment Setup and Baselines.}
We compare \method on AirDialogue with two baselines. The first is AirConcierge, the previous SOTA on AirDialogue, which explicitly parses and executes SQL queries from the dialogue \citep{chen2020airconcierge}. The other is a standard language model (denoted as LM(GPT2-small)) trained on a dataset filtered for successful task examples, without any of our context-aware language modeling techniques (see Appendix \Secref{app:filter} for more details on dataset filtering). \method uses the fine-tuned GPT2-small model \citep{radford2019language} as the backbone of the policy and dynamics model. After learning the dynamics model, both \method and the LM(GPT2-small) can employ two different planning strategies: (1) a simple greedy decoding of the next utterance (equivalent to beam search with beam-width one) and (2) the rollout planning as described in \Secref{sec:rollouts}. For AirConcierge, we only evaluate greedy decoding, as this method cannot be easily adapted for producing full rollouts. Rollout planning requires a method for predicting the reward of a given dialogue, and we describe our specific reward predictor for AirDialogue in Appendix \Secref{app:rew}.

\paragraph{Results for Task Success.}

\begin{figure}
    \centering
\includegraphics[scale=0.3]{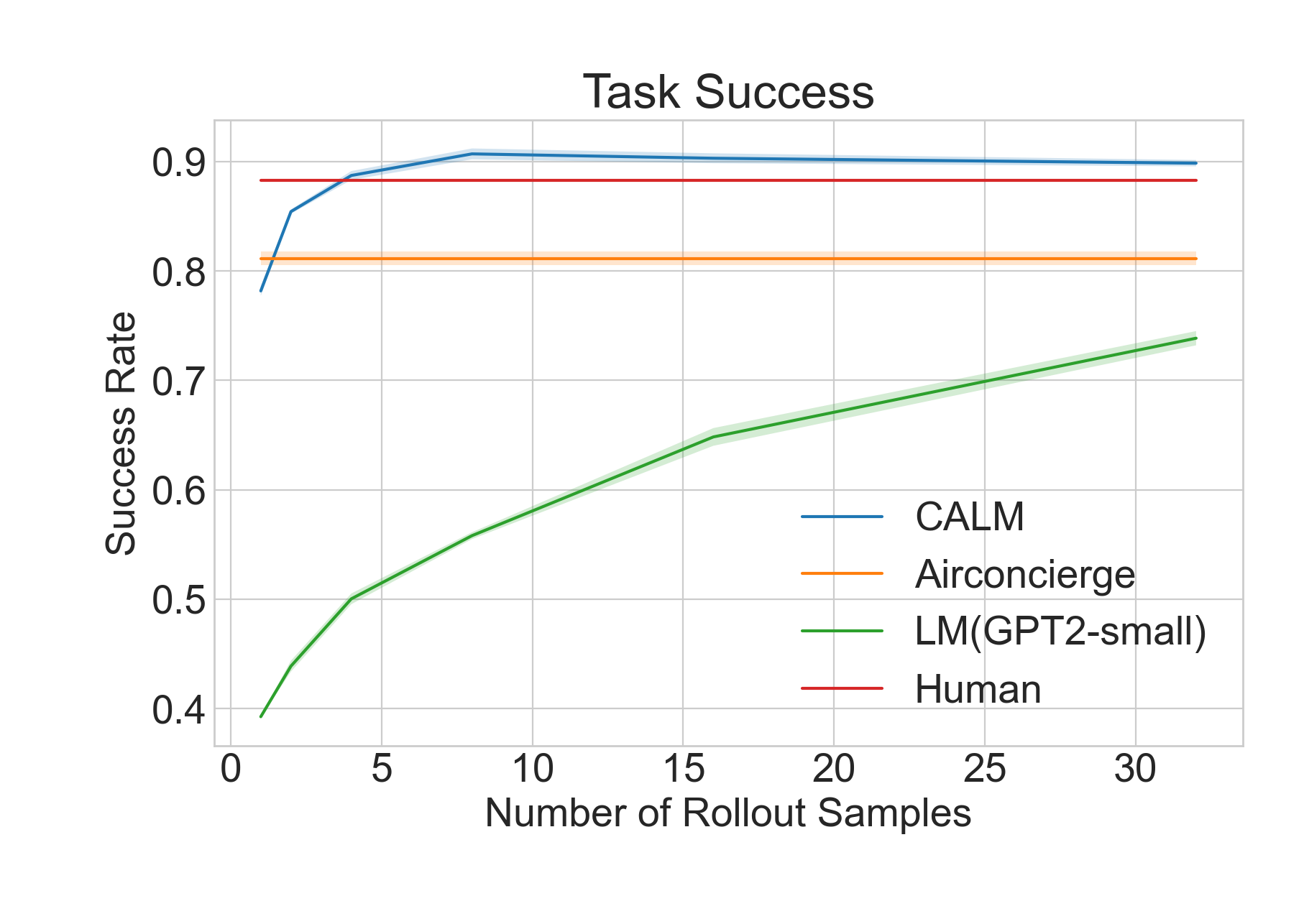}
    \vspace{-0.2in}
        \caption{\textbf{Task success as a function of the number of rollout samples.} Note that successful task completion improves with more rollout samples.}
    \label{fig:RolloutResults}
\end{figure}

In terms of task success, \method outperforms the prior SOTA (AirConcierge) by approximately 7\%, achieving 88\% task success when using greedy decoding from the language model (see Table \ref{tab:main_eval}). Compared with AirConcierge, where all reasoning about the task context is done outside of the language model, \method does all of the filtering, selecting, and responding with relevant flight table entries within the language model, in a fully end-to-end manner. Meanwhile, \method also improves over LM(GPT2-small) by 50\% in terms of task success, indicating the necessity of our context-aware approach for goal-oriented tasks.

We further evaluate the the performance of various methods, when utilizing the rollout planning technique. As shown in Figure \ref{fig:RolloutResults}, as the number of rollout samples increases, the performance improves for all methods. Remarkably, applying the rollout planning to \method further increases total task success by 2\%, raising it to~90\% and matching human performance on the AirDialogue task. The baseline LM(GPT2-small) benefits much more from rollout planning than \method, and we suspect that at around 90\% task completion, the performance becomes bottlenecked by the customer bot's mistakes, therefore we only observe less gain from rollout planning with \method.  

\paragraph{Results for Language Quality.}
To quantitatively measure the generated language quality, we present perplexity and BLEU for all methods in Table \ref{tab:language_eval}. \method performs similarly to LM(GPT2-small) and outperforms AirConcierge significantly.

\begin{table}[]
    \footnotesize
    \centering
    \begin{tabular}{|c|c|c|c|}
        \hline
        & \method & LM(GPT2-small) &  AirConcierge\\
        \hline
        Perplexity & 1.63 & \textbf{1.59} & - \\
        BLEU & 32.88 & \textbf{35.75} & 27.75 \\
        \hline
    \end{tabular}
    \caption{\textbf{BLEU score and perplexity results.} \method improves on task success without sacrificing generation quality.}
    \label{tab:language_eval}
\end{table}

\paragraph{Ablation Study.}
To examine the effectiveness of each single component in our method, 
we train and evaluate four ablations of \method. Each of these ablations remove one of the components in our approach: task relabeling (\Secref{sec:relabel}), auxiliary loss (\Secref{sec:auxiliary}), and table pre-training (\Secref{sec:pretrain}).
Beyond this, we also examine \method without both task relabeling and pre-training. As shown in Table~\ref{tab:ablation}, removing any one of these components drops task success by at least 10\%, and in most cases much more than that. This shows that each piece of our method plays a critical role in helping \method to effectively learn the goal-oriented task.

\begin{table}[]
    \footnotesize
    \centering
    \begin{tabular}{|l|c|}
        \hline
        & Success Rate \\
        \hline
        \method & \textbf{0.88} $\pm$ 4e-3 \\
        LM(GPT2-small) & 0.38 $\pm$ 1e-3 \\
        \method w/o relabel, pre-train & 0.42 $\pm$ 4e-3 \\
        \method w/o relabel & 0.66 $\pm$ 1e-2 \\
        \method w/o pre-train & 0.39 $\pm$ 3e-3 \\
        \method w/o auxiliary loss & 0.78 $\pm$ 4e-3\\
        \hline
    \end{tabular}
    \caption{\textbf{Task success rate for various ablations of \method on AirDialogue (all using greedy decoding).} Removing any single component from \method drops performance by at least 10\%.}
    \label{tab:ablation}
\end{table}
\section{Conclusion}

We proposed an end-to-end framework, \method, for goal-oriented dialogue systems. Formulating end-to-end dialogue generation as a Markov decision process, \method employs task relabeling and context-aware finetuning to steer supervised learning of language models towards specific goals, improving task performance drastically while preserving language quality. We show that this improves performance on AirDialogue over the previous state of the art, and matches previously reported human performance under the standard simulated evaluation protocol.

CALM optimizes for task-specific measures of success, and while such measures might be comparatively simple for domains such as AirDialogue, in general specifying the right success measure or reward function may present challenges. Furthermore, as with all methods based on end-to-end language models, CALM is susceptible to internal biases and inconsistencies in the language model itself. There is for example no constraint that ensures that CALM produces \emph{truthful} answers, or that it avoids harmful or socially unacceptable outputs. A practical deployable dialogue system would likely require additional measures to account for such issues, analogously to how learning-based methods for self-driving vehicles might require some additional safety mechanisms to ensure constraints, and indeed further research on reward specification, ensuring truthful outputs, and other constraint strategies for dialogue systems that combine language models and reward maximization is a promising and important direction.

The context-conditioned supervised learning strategy used by CALM provides for reward maximization, but is in general not optimal for arbitrary reinforcement learning problems: in general RL settings, learning a value function with dynamic programming in general can attain significantly better returns than imitating high-performing trajectories, by recombining good parts of multiple different trajectories (which might individually be suboptimal)~\citep{kostrikov2021offline,kumar2022should}. The simple supervised learning strategy works well in the domain we tested, but extending CALM to use value-based reinforcement learning methods is a promising direction for future work. Indeed, the improvement obtained from planning on top of the CALM model likely indicates that the supervised learning approach we employ has room for improvement. Additionally, the auxiliary objectives and relabeling strategies we employ require some amount of domain-specific design, and more general strategies could be developed in future.

Addressing these limitations in future work and developing more advanced methods that combine end-to-end language generation via large language models with concepts from reinforcement learning and planning is a promising research direction for making dialogue systems more capable, while also making language models more task aware. We hope that CALM will serve as an indication for the potential of such methods.

\section*{Acknowledgements}

This research was supported by an Amazon-BAIR Commons project, as well as the Office of Naval Research. We thank Dilek Hakkani-tur, Alexandros Papangelis, Mandi Zhao, Ruiqi Zhong, and Yang Liu for advice and feedback.

\bibliographystyle{acl_natbib}
\bibliography{acl}
\pagebreak
\appendix
\clearpage

\section{AirDialogue Dataset Filtering}
\label{app:filter}
When training the LM(GPT2-small) and Customer Bot, we filter the dataset by only keeping the successful task examples. This is be achieved by simultaneously checking for successful task completion and whether a set of simple string matching heuristics are satisfied in the dialogue. Our heuristics aim to ensure that strings corresponding to each of the customer's flight requirements and the customer's goal are explicitly present in the dialogue. This combination of filtering steps reduces the size of the training set by 26\%. Despite this, we find that this is still more than enough data for the model to successfully learn the task.

\subsection{Rollout Planning}
\begin{figure}[h]
    \centering
    \includegraphics[scale=0.2]{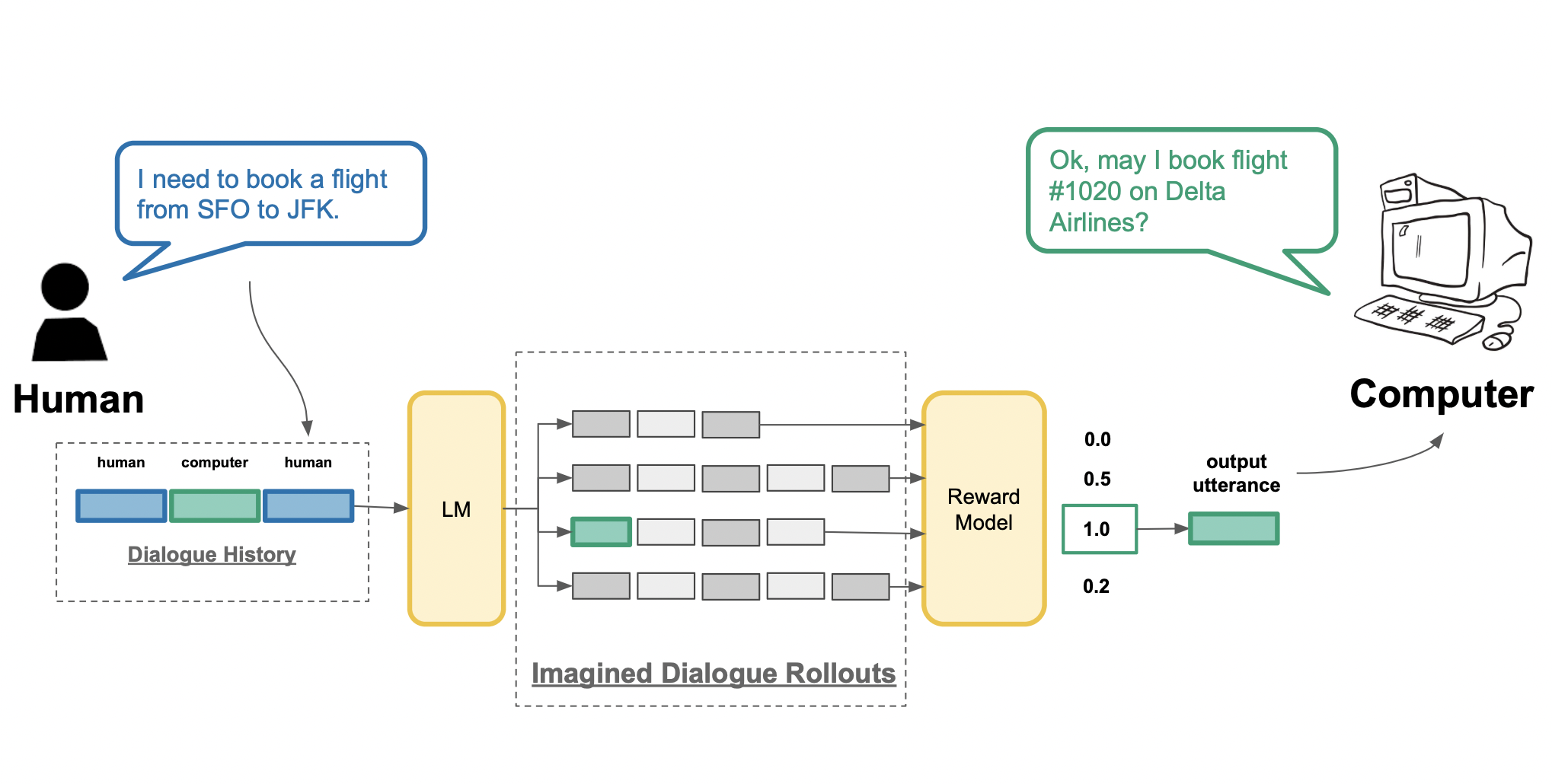}
    \caption{\textbf{Our dialogue rollout planning procedure.} To generate our response, we sample entire dialogues from the language model and then re-rank the predicted dialogues with a reward function.}
    \label{fig:MCTS}
\end{figure}

In Figure~\ref{fig:MCTS}, we show the rollout planning procedure, which described in Section~\ref{sec:rollouts}.

\section{Training Our Customer Bot}
\label{app:cus-bot}
Our customer bot is fine-tuned from GPT2-small (124M parameters), using the standard language modeling objective. We used the Huggingface Transformers library's implementation of GPT2 \citep{wolf2020huggingfaces}. The customer's flight requirements are provided to the model as a prefix to the dialogue, which formatted as a comma separated list consisting of the customer's goal and flight requirements.
We trained the customer bot for maximum 10 epochs with early stopping on the filtered dataset. For training, it takes around 1 day on 4 GPUs. Specifically, we trained using Adam with learning rate 1e-4 and batch size 8. 
Our customer bot achieves a perplexity of 1.47 on the development set and a BLEU score of 38.5.

\section{Fight Agent Bot Details}
\label{app:fl-bot}
All our flight agent bots are fine-tuned from GPT2-small (124M parameters) using the standard language modeling objective. We used the Huggingface Transformers library's implementation of GPT2 \citep{wolf2020huggingfaces}. Similar as the customer bot, we trained for maximum 10 epochs with early stopping on the filtered dataset, which takes roughly 1 day on 4 GPUs. Specifically, we trained using Adam with learning rate 1e-4 and batch size 8. We implement the final action prediction as a sequence of tokens generated at the end of each dialogue. The flight table is passed to the model as a prefix of flight embeddings, where each embedding is produced by summing embeddings corresponding to each attribute of a given flight (e.g., flight arrival/departure day/location, flight price, etc.).

\section{AirDialogue Task Pretraining}
\label{app:pretrain}
Initialized using GPT2-small (124M parameters), we further pre-train our flight-agent bots by training on simplified task sequences. Specifically, these sequences consist of our flight table followed by a comma separated list of the customer's flight requirements and a string representing the final action taken. We also apply our auxiliary loss and task-relabeling techniques during this pre-training.

We pre-train on 4 million unique samples, using batch size 64 and Adam with learning rate 1e-4, which takes around 2 days on 4 GPUs. During pre-training, we found that it took around $2$ million unique samples before the model suddenly started to learn the task of querying the flight table, and it took roughly $2$ million more samples before it became proficient at querying the table. Both the unusual progression of learning during this pre-training phase and the high sample complexity needed to learn the task, indicates the difficulty in learning to query the flight table. This calls for future work about further investigate the challenges in learning complex logical functions using neural networks.

\section{Self-Play Evaluation}
\label{app:self-play}
Prior works primarily evaluate bots for the flight agent through ``self-play" \citep{chen2020airconcierge, wei2018airdialogue}. We follow the same evaluation protocol in our work. Basically, we train a bot to play the role of the customer during evaluation and compute task success by simulating conversations against this bot. We run all self-play evaluations on the same subset of 1,000 dialogue scenarios, randomly selected from the validation set. 

All models are evaluated against the same customer bot. including models for the baselines. We find that when running against our self-play bot, task completion success for prior methods is increased, sometimes by more than 8\% (from what was reported by such prior works under the same evaluation setting). The only difference is the specific model used for customer's side of the conversation, and we conjecture that this difference is likely due to the architecture difference and the details of our dataset filtering. This significant change in evaluation performance compared with prior works, not only indicates the quality of our customer bot, but also suggests the importance of accounting for these factors in evaluating and comparing dialogue systems. We release the code and model weights for our customer bot at \url{https://sea-snell.github.io/CALM_LM_site/}.

\subsection{AirDialogue Reward Predictor for Rollout Planning}
\label{app:rew}
\begin{figure}
    \centering
    \includegraphics[scale=0.2]{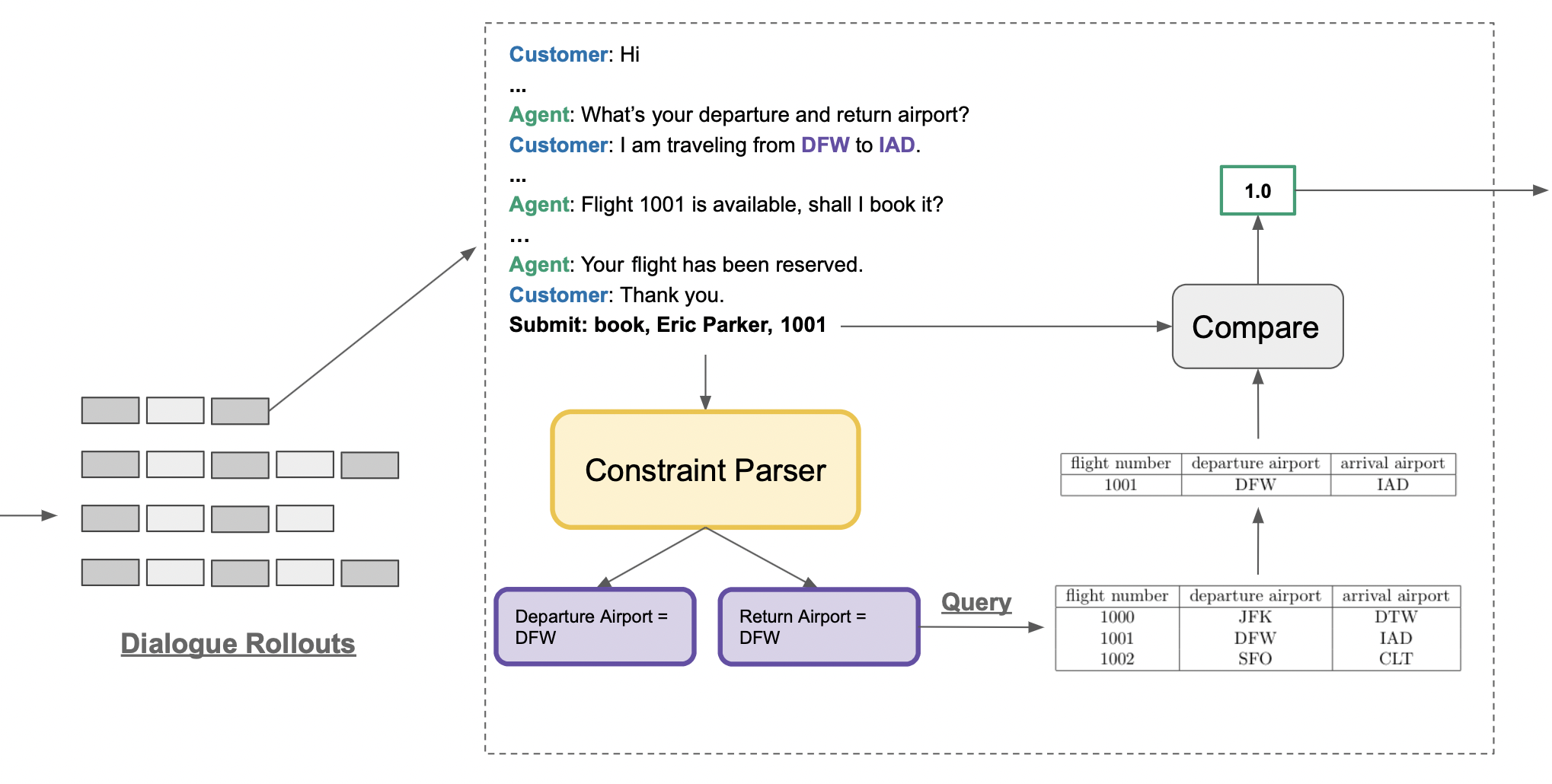}
    \caption{\textbf{Our reward prediction method.} We train a model to parse the customer's flight requirements from the dialogue. We execute these flight requirements against the table and compare the output to the flight that was actually booked; this determines the reward (i.e. if the correct flight was booked or not).}
    \label{fig:rewardpredict}
\end{figure}

To execute rollout planning, we need a reward predictor which can estimate whether a given dialogue is a successful example of task completion or not. In the case of AirDialogue, we found that the most robust way to estimate this reward is the following: we first fine-tune a RoBERTa-base model (123M parameters) to predict the customer's ground-truth goal and flight requirements from the set of dialogues in the training set. We used the Huggingface Transformers library's implementation of RoBERTa \citep{wolf2020huggingfaces}. We do not filter the training-set when training this model. Once this model is trained, our procedure for predicting dialogue success is the following:
\begin{enumerate}
    \item Given a dialogue, use our RoBERTa model to predict the customer's goal and flight requirements.
    \item We then execute this predicted information against the agent's flight table and reservation flag, to produce a set of valid final actions.
    \item If the final action taken in the dialogue is within the set of predicted final actions, then predict that the current dialogue is successful, otherwise predict that it is unsuccessful.
\end{enumerate}
See Figure \ref{fig:rewardpredict} for a visual illustration of this procedure. Our model obtains 94\% accuracy in predicting the reward of the dialogues in the validation set (see Table \ref{tab:ConstraintPredictor} for a more extensive breakdown of the model's accuracy).

\begin{table}[]
    \centering
    \begin{tabular}{|c|c|c|c|c|}
         \hline
         dep. city & ret. city & dep. month & ret. month \\
         \hline
         0.76 & 0.76 & 0.77 & 0.77 \\
         \hline
         \hline
         dep. day & ret. day & dep. time & ret. time \\
         \hline
         0.76 & 0.76 & 0.94 & 0.94 \\
         \hline
         \hline
         class & price & connections & airline \\
         \hline
         0.92 & 0.37 & 0.95 & 0.97 \\
         \hline
    \end{tabular}
    \caption{\textbf{Our RoBERTa parser's accuracy in predicting each of the customer's flight requirements.} The parser predicts 5 out of 12 flight requirements with >90\% accuracy and 11 out of 12 with >70\% accuracy. The price requirement has the lowest accuracy because it is often not explicitly mentioned in the dialogue; the model has to rely on priors for prediction in these cases.}
    \label{tab:ConstraintPredictor}
\end{table}

\section{Example Conversation in AirDialogue}
\begin{figure}[h]
    \centering
    \includegraphics[scale=0.3]{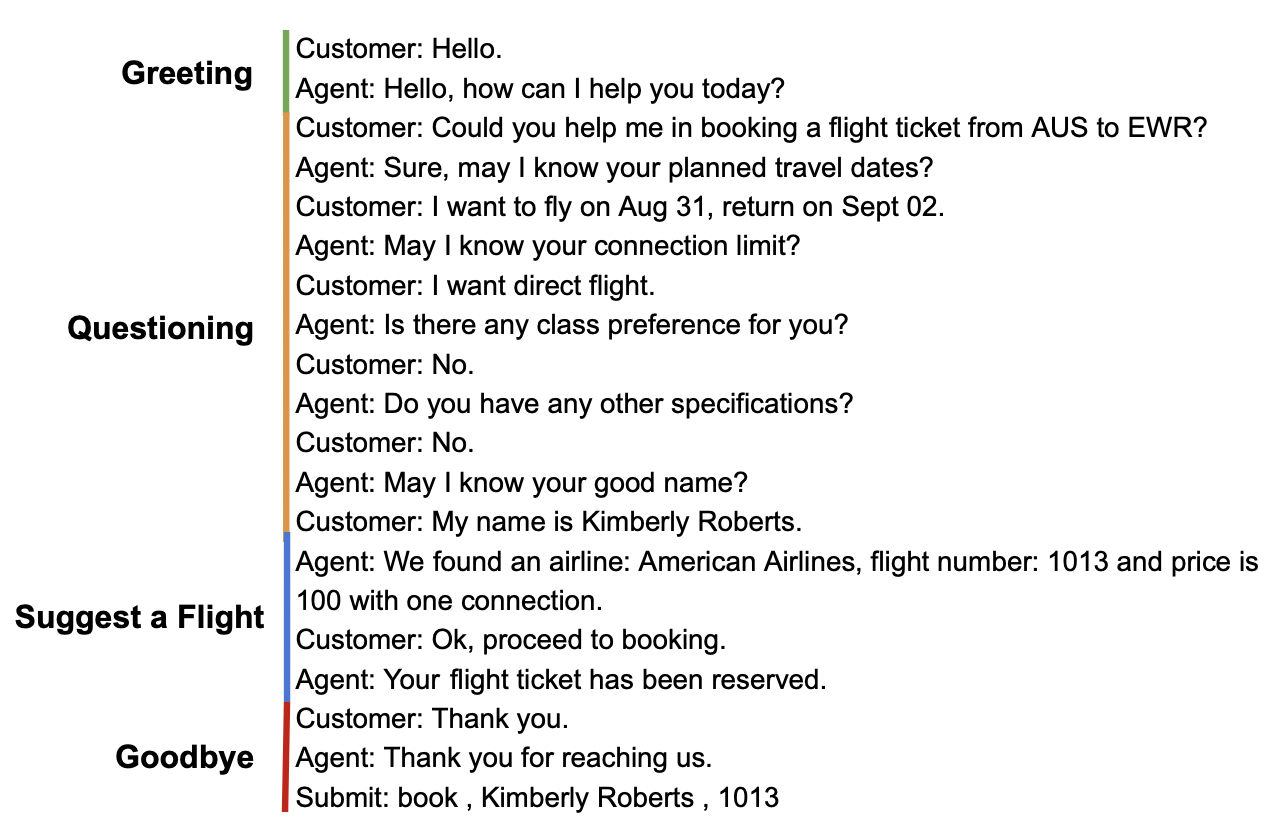}
    \caption{\textbf{An example conversation in AirDialogue.} Conversations generally begin with a greeting followed by some questioning / information gathering, and then finally the agent suggests a flight before ending the conversation.}
    \label{fig:AirDialogueExample}
\end{figure}

In Figure~\ref{fig:AirDialogueExample}, we showcase a specific example for the conversation in AirDialogue.

\section{Previous Approaches to Flight Table Processing}
\label{app:table}
Prior works \citep{wei2018airdialogue, jiang2021towards} typically input the table directly into a language model, expecting that the skill of querying the table will be naturally learned via the standard language modeling objective. We found this approach to under-perform in our experiments. These findings are also consistent with recent works which show that pre-training transformers for querying tables can significantly improve the transformer's performance on downstream tasks which use tables \citep{liu2021tapex}. AirConcierge \citep{chen2020airconcierge} takes a different approach, and explicitly predicts and executes SQL queries based on the dialogue. This approach obtains the SOTA task success on AirDialogue, but it involves several complex components, requires the ability to preform semantic parsing on the dialogue, and of course requires additional domain knowledge about the format and structure of the flight table, which reprsents the task context. In our work, we show that applying \method for AirDialogue can close this gap by inducing task learning from language models and achieve end-to-end learning from the flight table, without sacrificing the generated language quality.

\begin{table*}[]
    \centering
    \begin{tabular}{|r|c|c|c|}
         \hline
          & \method & LM (GPT2-small) & AirConcierge \\
         \hline
         full success rate & \textbf{0.88}$\pm$4e-3 & 0.38$\pm$1e-3 & 0.81$\pm$7e-3 \\
         \hline
         status success rate & \textbf{0.92}$\pm$3e-3 & 0.84$\pm$2e-3 & 0.90$\pm$1e-3 \\
         \hline
         flight success rate & \textbf{0.88}$\pm$4e-3 & 0.39$\pm$1e-3 & 0.82$\pm$5e-3 \\
         \hline
         name accuracy rate & 0.99$\pm$2e-3 & \textbf{1.0}$\pm$8e-4 & 0.99$\pm$1e-3 \\
         \hline
         \multirow{3}{*}{book R/P/F1} & \textbf{0.85}$\pm$8e-3 & 0.06$\pm$3e-3 & 0.81$\pm$1e-2 \\ 
                                      & \textbf{0.86}$\pm$8e-3 & 0.05$\pm$2e-3 & 0.70$\pm$9e-3 \\ 
                                      & \textbf{0.85}$\pm$6e-3 & 0.05$\pm$3e-3 & 0.75$\pm$1e-2 \\
         \hline
         \multirow{3}{*}{no flight R/P/F1} & \textbf{0.82}$\pm$1e-2 & 0.36$\pm$9e-3 & 0.59$\pm$5e-3 \\
                                           & 0.80$\pm$1e-2 & 0.74$\pm$5e-3 & \textbf{0.93}$\pm$6e-3 \\ 
                                           & \textbf{0.81}$\pm$1e-3 & 0.49$\pm$9e-3 & 0.72$\pm$5e-3 \\
         \hline
          \multirow{3}{*}{cancel R/P/F1} & 0.98$\pm$2e-2 & \textbf{1.0}$\pm$0.0 & \textbf{1.0}$\pm$0.0 \\
                                         & 0.95$\pm$3e-2 & \textbf{1.0}$\pm$0.0 & 0.75$\pm$1e-2 \\
                                         & 0.97$\pm$2e-2 & \textbf{1.0}$\pm$0.0 & 0.86$\pm$7e-3 \\
         \hline
          \multirow{3}{*}{change R/P/F1} & \textbf{0.25}$\pm$8e-2 & 0.0$\pm$0.0 & 0.0$\pm$0.0 \\
                                         & \textbf{0.33}$\pm$1e-1 & 0.0$\pm$0.0 & 0.0$\pm$0.0 \\
                                         & \textbf{0.28}$\pm$1e-1 & 0.0$\pm$0.0 & 0.0$\pm$0.0 \\
         \hline
          \multirow{3}{*}{no reservation R/P/F1} & \textbf{0.99}$\pm$5e-3 & \textbf{0.99}$\pm$2e-3 & \textbf{0.99}$\pm$3e-3 \\
                                                 & \textbf{0.99}$\pm$2e-3 & \textbf{0.99}$\pm$2e-3 & \textbf{0.99}$\pm$3e-3 \\
                                                 & \textbf{0.99}$\pm$3e-3 & \textbf{0.99}$\pm$2e-3 & \textbf{0.99}$\pm$3e-3 \\
         \hline
         constraint success & 0.81$\pm$9e-3 & 0.71$\pm$3e-3 & \textbf{0.89}$\pm$1e-3 \\
         \hline
    \end{tabular}
    \caption{\textbf{Detailed statistics for model errors.} All models are evaluated with greedy decoding. In addition to the full task success rate, we report success rate for each sub-component of the full task (status / flight / name). We also report recall (R), precision (P), and F1 score for task success under each type of high-level action (book / no flight / cancel / change / no reservation). Lastly, we report the average fraction of the customer's flight requirements that are met when the agent books the wrong flight (constraint success).}
    \label{tab:ErrorAnalysis}
\end{table*}

\section{Error Analysis}

In Table \ref{tab:ErrorAnalysis} we present a detailed breakdown of model errors. As expected, determining the flight to book, if any, is consistently shown to be the most challenging sub-task, as evidenced by the lower ``flight success rate" and the lower F1 scores for ``no flight", ``book", and ``change" on LM (GPT2-small). In particular, ``change" has a low recall, precision, and F1 score for all models because it makes up a very small 0.4\% of the training data. Lastly, the ``constraint success" row shows that even when \method books the wrong flight, the flight it does books meets >80\% of the customer's flight requirements on average.

\end{document}